\documentclass[10pt,twocolumn,letterpaper]{article}

\usepackage{cvpr}
\usepackage{times}
\usepackage{epsfig}
\usepackage{graphicx}
\usepackage{amsmath}
\usepackage{amssymb}
\usepackage{animate}
\usepackage{enumitem}

\usepackage[pagebackref=true,breaklinks=true,letterpaper=true,colorlinks,bookmarks=false]{hyperref}

\cvprfinalcopy 

\def\cvprPaperID{591} 

\ifcvprfinal\fi
 
\newcommand{\ATT}[1]{{\color{black} #1}} 
\newcommand{\subsup}[3]{{#1}_{\mkern-4mu #2}^{\mkern-4mu #3}}
\begin{document}

\title{Video2GIF: Automatic Generation of Animated GIFs from Video}

\author{Michael Gygli\thanks{This work was done while the author was an intern at Yahoo! Inc.}\\
CVL, ETH Zurich\\
Zurich, Switzerland\\
{\tt\small gygli@vision.ee.ethz.ch}
\and
Yale Song\\
Yahoo Research\\
New York, USA\\
{\tt\small yalesong@yahoo-inc.com}
\and
Liangliang Cao\\
Yahoo Research\\
New York, USA\\
{\tt\small liangliang@yahoo-inc.com}
}

\maketitle

\begin{abstract}
\vspace{-2mm}
We introduce the novel problem of automatically generating animated GIFs from video. GIFs are short looping video with no sound, and a perfect combination between image and video that really capture our attention. GIFs tell a story, express emotion, turn events into humorous moments, and are the new wave of photojournalism. We pose the question: Can we automate the entirely manual and elaborate process of GIF creation by leveraging the plethora of user generated GIF content? We propose a Robust Deep RankNet that, given a video, generates a ranked list of its segments according to their suitability as GIF. We train our model to learn what visual content is often selected for GIFs by using over 100K user generated GIFs and their corresponding video sources. We effectively deal with the noisy web data by proposing a novel adaptive Huber loss in the ranking formulation. We show that our approach is robust to outliers and picks up several patterns that are frequently present in popular animated GIFs. On our new large-scale benchmark dataset, we show the advantage of our approach over several state-of-the-art methods.
\end{abstract}

\vspace{-3mm}
\section{Introduction}
Animated GIF is an image format that continuously displays multiple frames in a loop, with no sound. Although first introduced in the late 80's, its popularity has increased dramatically in recent years on social networks, such as Tumblr and reddit, generating numerous famous Internet memes and creative Cinemagraphs~\cite{BSKSJK2016}. In response, various websites have been created to provide easy-to-use tools to generate GIF from video, e.g., GIFSoup, Imgflip, and Ezgif. However, while becoming more prevalent, the creation of GIF remains an entirely manual process, requiring the user to specify the timestamps of the beginning and the end of a video clip, from which a single animated GIF is generated. This way of manually specifying the exact time range makes existing solutions cumbersome to use and requires extensive human effort. 
\begin{figure}
\centering
\includegraphics[clip, trim=15 0 20 0,width=1.0\linewidth]{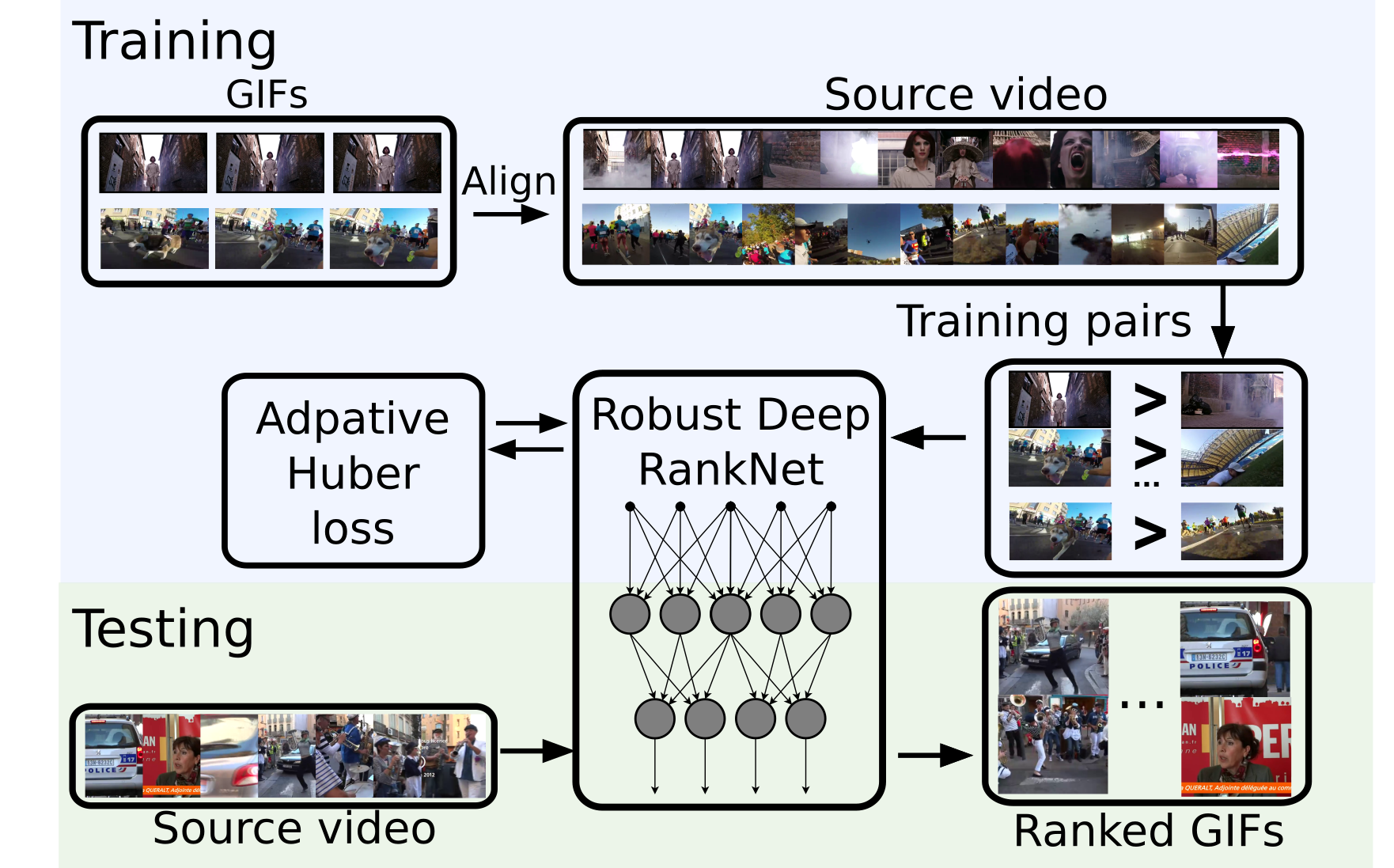}
\caption{Our goal is to rank video segments according to their suitability as animated GIF. We collect a large-scale dataset of animated GIFs and the corresponding video sources. This allows us to train our Robust Deep RankNet using over 500K pairs of GIF and non-GIF segment pairs, learning subtle differences between video segments using our novel adaptive rank Huber loss. 
}
\label{fig:pipeline}
\end{figure}

In this paper, we introduce the novel problem of automatically generating animated GIFs from video, dubbed Video2GIF. From the computer vision perspective, this is an interesting research problem because GIFs have some unique properties compared to conventional images and videos: A GIF is short, entirely visual with no sound, expresses various forms of emotions,
and sometimes contains unique spatio-temporal visual patterns that make it appear to loop forever. The task has some connections to existing computer vision problems -- such as visual interestingness~\cite{Jiang2013,Gygli2013}, creativity~\cite{Redi}, video highlights~\cite{Sun2014,Yang2015} and summarization~\cite{Gygli2015,Song} -- but differs from them due to the unique properties described above. Apart from research interest, the task is supported by real-world demand and has many practical application scenarios including photojournalism, advertising, video sharing and preview, as well as video promotion on social media.

To handle this task, we propose a novel RankNet that, given a video, produces a ranked list of segments according to their suitability as animated GIF. Our framework has several novel components designed to learn what content is frequently selected for popular animated GIFs. First, to capture the highly dynamic spatio-temporal visual characteristics of GIFs, we use 3D convolutional neural networks~\cite{Tran2014a} to represent each segment of a video. Second, to unravel the complex relationships and learn subtle differences between segments of a given video, we construct a ranking model that learns to compare pairs of segments and find the ones that are more suitable as GIF. Third, to make our learning task robust to the noisy web data, we design a new robust adaptive Huber loss function in the ranking formulation. Lastly, to account for different degrees of quality in user generated content, we encode the popularity measure of GIFs on social media directly into our loss.

Crucial to the success of our approach is our new large-scale animated GIF dataset: We collected more than 100K user generated animated GIFs with their corresponding video sources from online sources. There are hundreds of thousands of GIFs available online and many provide a link to the video source. This allows us to create a dataset that is one to two orders of magnitude larger than existing datasets in the video highlight detection and summarization literature~\cite{Sun2014,Song,Gygli2014}. We use this dataset to train our deep neural network by making comparisons between more than 500K GIF and non-GIF pairs. Experimental results suggest that our model successfully learns what content is suitable for GIFs, and that our model generalizes well to other tasks, namely video highlight detection~\cite{Sun2014}.

In summary, we make the following contributions:
\begin{enumerate}[nolistsep]
\item{We introduce the task of automatically generating animated GIFs from video. This is an interesting computer vision research problem that, to the best of our knowledge, has not been addressed before.}
\item{We propose a Robust Deep RankNet with a novel adaptive Huber loss in the ranking formulation. We show how well our loss deals with noisy web data, and how it encodes the notion of content popularity to account for different degrees of content quality.}
\item{We collect a new, large-scale benchmark dataset of over 100K user generated animated GIFs and their video sources. The dataset is one to two orders of magnitude larger than existing video highlighting and summarization datasets. The dataset is publicly available\footnote{\url{https://github.com/gyglim/video2gif_dataset}}.}
\end{enumerate}

\section{Related Work}
Our work is closely related to image aesthetics and interestingness, as well as video highlight detection and summarization. We review some of the most relevant work and discuss the differences. We also review and make connections to recent efforts on learning deep neural networks for ranking and trained on large-scale weakly-labeled data.

\smallskip
\noindent
\textbf{Image aesthetics and interestingness.}
Finding the best images in a collection has been studied from several angles. Early approaches aimed at predicting the quality~\cite{Ke2006} or aesthetics~\cite{Datta,Dhar2011} of an image. More recently, several approaches for predicting visual interestingness of an image have been proposed~\cite{Grabner2013,Gygli2013,Fu2014}. While interestingness is a subjective property assessed by the viewer, there is considerable consistency across annotated ratings~\cite{Gygli2013}. This makes it possible to model interestingness with computational means, but ground truth is typically noisy. Fu~\etal~\cite{Fu2014} propose an approach accounting for this, by learning a ranking model and removing outliers in a joint formulation. Khosla~\etal\cite{Khosla} analyze the related property of image popularity.
Using a large-scale dataset of Flickr images, they analyze and predict what types of images are more popular than others, surfacing trends similar to those of interestingness~\cite{Gygli2013}. In a similar direction is the work of Redi~\etal~\cite{Redi}, which analyzes creativity. Rather than analyzing images, however, they focus on Vines videos, whose lengths are restricted to 6 seconds.

\smallskip
\noindent
\textbf{Video summarization.}
A thorough discussion of earlier research can be found in~\cite{Truong2007}. Here, we discuss two recent trends, (i) using web-image priors~\cite{Kim,Khosla,Song,Liu2015} and (ii) supervised learning-based methods~\cite{Ghosh2012,Gonga,Gygli2015,Liu2015}. Methods using web-image priors are based on the observation that web images for a specific topic or query are often canonical visual examples for the topic. This allows one to compute frame scores as the similarity between a frame and a set of web images~\cite{Kim,Khosla,Song}. Learning-based methods, on the other hand, use supervised models to obtain a scoring function for frames~\cite{Ghosh2012,Gonga,Liu2015} or segments~\cite{Gygli2015}. Lee~\etal~\cite{Ghosh2012} learn a regression model and combine it with a clustering approach to diversify the results. Instead,~\cite{Gonga,Gygli2015} directly learn an objective function that scores a set of segments, based on relative importance between different aspects of a summary (e.g. balancing highlights and diversity). Crucial to these learning-based methods is some notion of importance or interestingness of a segment. Next, we will discuss methods focusing only on this part while ignoring diversity and information coverage.

\smallskip
\noindent
\textbf{Video highlights.}
The definition of highlight is both subjective and context-dependent~\cite{Truong2007}. Nevertheless, it has been shown that there exists some consistency among human ratings for this task~\cite{Song,Gygli2014}. Several methods exploit, for example, that close-ups of faces are generally of interest~\cite{Truong2007,Ghosh2012,Gygli2014}. But these approaches are limited in that they rely on a few hand-crafted features for capturing highlights in highly diverse settings. Instead, several approaches for domain-specific models have been proposed. In particular, in sport games highlight is more clearly defined (\eg scoring a goal) which has been exploited in many works (see~\cite{Truong2007} for an overview). Recently, Sun~\etal~\cite{Sun2014} and Potapov~\etal~\cite{Potapov2014} proposed a more general approach. Based on annotated videos for a specific topic (\eg surfing), they use machine learning on top of generic features to train a highlight predictor. In order to train their model,~\cite{Potapov2014} uses a large, manually annotated dataset for action recognition. Instead,~\cite{Sun2014} use a smaller dataset obtained by crawling YouTube data. They find pairs of raw and edited videos, used in training, by matching all pairs of videos within a certain category (\eg gymnastics). The size of their dataset is, however, limited by the availability of domain-specific videos in both raw and edited forms.

Obtaining a large-scale video highlight dataset is difficult. Thus, Yang et al.~\cite{Yang2015} propose an unsupervised approach for finding highlights. Relying on an assumption that highlights of an event category are more frequently captured in short videos than non-highlights, they train an auto-encoder. Our work instead follows a supervised approach, introducing a new way to obtain hundreds of thousands of labeled 
training videos (10x larger than the unlabeled dataset of~\cite{Yang2015}), which allows us to train a deep neural network with millions of parameters.

\smallskip
\noindent
\textbf{Learning to rank with deep neural networks.}
Several works have used CNNs to learn from ranking labels.
The loss function is often formulated over pairs~\cite{Liu2015,Gong2013} or triplets~\cite{Wang2014a,Wang2015,Zhao2015,Lai2015}.
Pairwise approaches typically use a single CNN, while the loss is defined relatively over the output. For example, Gong~\etal~\cite{Gong2013} learn a network to predict image labels and require the scores of correct labels to be higher than the scores of incorrect labels.
Triplet approaches, on the other hand, use Siamese networks. Given an image triple (query, positive, negative), a loss function requires the learned representation of the query image to be closer to that of the positive, rather than the negative image, according to some metric~\cite{Wang2014a,Wang2015,Zhao2015,Lai2015}.

\smallskip
\noindent
\textbf{Supervised deep learning from noisy labels.}
Several previous works have successfully learned models from weak labels~\cite{Karpathy2014,Wang2014a,Liu2015}.
Liu~\etal~\cite{Liu2015} considers the video search scenario.
Given click-through data from Bing, they learn a joint embedding between query text and video thumbnails in order to find semantically relevant video frames.
In contrast,~\cite{Karpathy2014,Wang2014a} use labels obtained through automatic methods to train neural networks.
Karpathy~\etal~\cite{Karpathy2014} train a convolutional neural network for action classification in videos. 
Their training data is obtained from YouTube where it is labeled automatically by analyzing meta data associated with the videos.
Wang~\etal~\cite{Wang2014a} learn a feature representation for fine-grained image ranking. 
Based on existing image features they generate labels used for training the neural network.
Both approaches obtain state-of-the-art performance, showing the strength of large, weakly-labeled datasets in combination with deep learning.

\section{Video2GIF Dataset}
\label{sec:dataset}

Inspired by the recent success with large, weakly-labeled datasets applied in combination with deep learning, we harvest social media data with noisy, human generated annotations. We use websites that allow users to create GIFs from video (Make-a-GIF and GIFSoup). Compared to edited videos used in~\cite{Sun2014}, GIFs have the intriguing property that they are inherently short and focused. Furthermore they exist in large quantities and typically come with reference to the initial video, which makes alignment scale linearly in the number of GIFs. Aligning GIFs to their source videos is crucial, as it allows us to find non-selected segments, which serve as negative samples in training. In addition, videos provide a higher frame-rate and fewer compression artifacts, ideal for obtaining high quality feature representations.

Using these GIF websites, we collected a large-scale dataset with more than \ATT{120K} animated GIFs and more than \ATT{80K} videos, with a total duration of \ATT{7,379} hours. This is one to two orders of magnitude larger than the highlight datasets of~\cite{Yang2015} and~\cite{Sun2014}. We will show further statistics on the dataset after discussing the alignment process.



\begin{table}
\centering
\begin{tabular}{|l|l|}
\hline
\textbf{Property}  & \textbf{Quantity} \\
\hline

Total number of animated GIFs  & 121,647 \\ 
Mean GIF duration & 5.8 sec \\ 

\hline \hline 
Total number of videos & 84,754 \\ 
Total video duration & 7,379 hr \\ 
Mean video duration & 5.2 min \\ 
Total number of videos (CC-BY) & 432\\ 

\hline \hline 
GIFs used in experiment & 100,699 \\ 
Videos used in experiment & 70,456 \\

\hline
\end{tabular}
\caption{Statistics on the Video2GIF dataset. We show numbers for the complete dataset and for the one after discarding too short or too long videos (see text). We also show the number of videos that come with the Creative Commons license (CC-BY).}
\label{tab:dataset}
\end{table}
\begin{figure}
\centering
\includegraphics[width=1.0\linewidth]{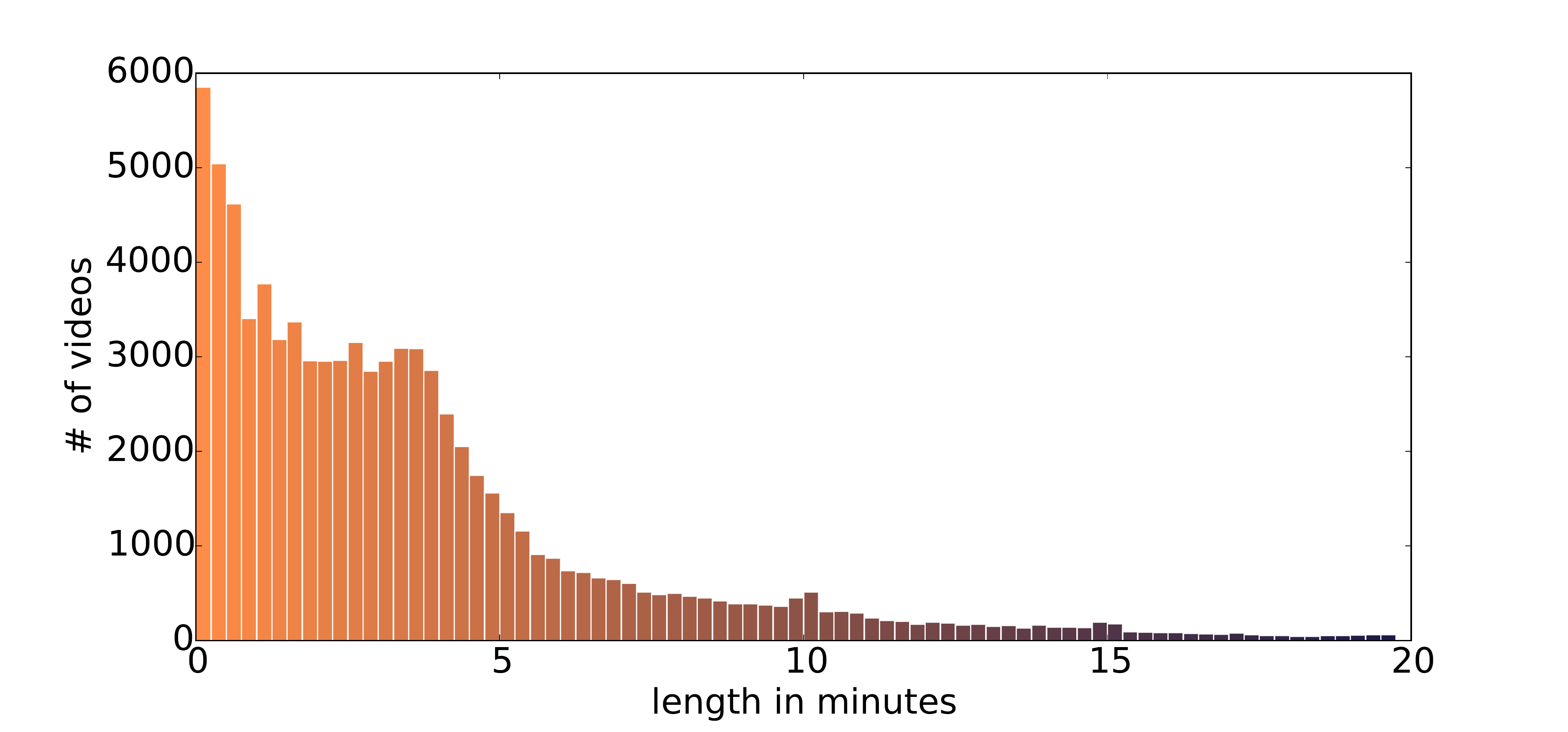}
\caption{Length distribution of the input videos.} 
\label{fig:video_length}
\end{figure} 

\begin{figure*}
\centering
\includegraphics[width=1.0\linewidth]{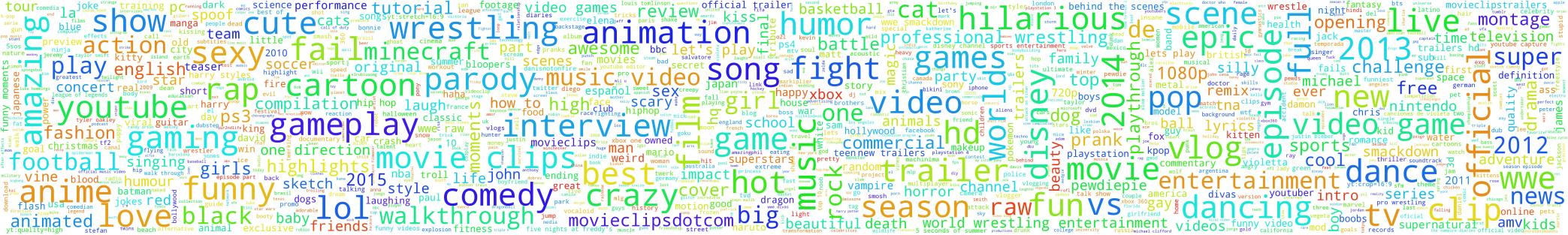}
\caption{Most frequent video tags on the used dataset. We can observe that not all tags are equally informative. 
While several describe a specific visual concept (~\eg cat or wrestling) others describe abstract concepts that cannot be expected to help the task at hand.}
\label{fig:tags}
\end{figure*}
\begin{figure*}
\centering
\includegraphics[clip,trim=20 95 20 0, width=1.0\linewidth]{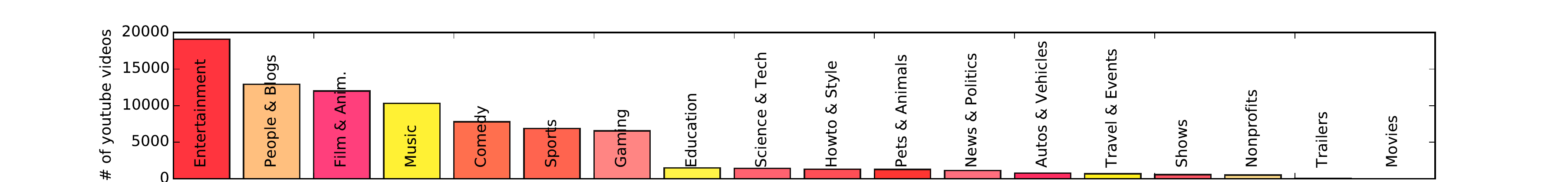} 
\caption{Distribution over video categories. Note how the categories are highly imbalanced and often not specific.
~\eg Entertainment is an extremely broad category with strong visual and semantic variation.}
\label{fig:categories}
\end{figure*}

\smallskip
\noindent
\textbf{Alignment.}
We aligned the GIFs to their corresponding videos using frame matching. In order to do this efficiently, we encoded each frame with a perceptual hash based on the discrete cosine transform~\cite{zauner2010implementation}. 
The perceptual hash is fast to compute and, given its binary representation, can be matched very efficiently using the Hamming distance.
We matched the set of GIF frames to the frames of its corresponding video. 
This approach requires $\mathcal{O}(nk)$ distance computations, where $n,k$ is the number of frames in the video and GIF, respectively.
Since the GIFs are restricted in length and have a low frame-rate, they typically contain only a few frames ($k<50$).
Thus, this method remains computationally efficient while it allows for the alignment to be accurate.

In order to test the accuracy of our alignment process, we manually annotated a small random set of 20 GIFs with ground-truth alignments and measured the error. Our method has a mean alignment error of $0.34$ seconds (median $0.20$ seconds), which is accurate enough for our purpose. 
In comparison, Sun~\etal~\cite{Sun2014} aligned blocks of 50 frames ($\approx 2$ seconds),~\ie on a much coarser level.

\smallskip
\noindent
\textbf{Dataset Analysis.}
We analyze what types of video are often used to create animated GIFs.
Figure~\ref{fig:tags} shows the most frequent tags of videos in our dataset, and Figure~\ref{fig:categories} shows the category distribution of the videos.
Several tags give a sense of what is present in the videos, which can potentially help GIF creation,~\eg cute and football. Others are not visually informative, such as 2014 or YouTube.
Figure~\ref{fig:video_length} shows a histogram of video lengths (median: 2m51s, mean: 5m12s). As can be seen, most source videos are rather short, with a median duration of less than 3 minutes. 


\smallskip
\noindent
\textbf{Splits.}
From the full dataset we used videos with a maximal length of 10 minutes. 
Longer videos are discarded as the selected GIF segments become too sparse and the videos are more affected by chronological bias~\cite{Song}.
We split the data into training and validation sets, with about \ATT{65K and 5K} videos, respectively. 
For the test set, we use videos with Creative Commons licence, which allows us to distribute the source videos for future research.
As the task is trivial for videos shorter than 30sec we only consider videos of longer duration. The final test set consists of \ATT{357} videos. 
Table~\ref{tab:dataset} shows the statistics of the dataset we used in our experiments.

\section{Method}
This sections presents our approach to the Video2GIF task, with a novel adaptive Huber loss in the ranking formulation to make the learning process robust to outliers; we call our model the Robust Deep RankNet.

\subsection{Video Processing}
We start by dividing a video into a set of non-overlapping segments $\mathcal{S} = \{ s_{1}, \cdots, s_{n} \}$. We use the efficient shot boundary detection algorithm of Song~\etal~\cite{Song}, which solves the multiple change point detection problem to detect shot boundaries. 

The segments are not necessarily aligned perfectly with the boundaries of the actual animated GIF segments. We determine whether a segment $s$ belongs to GIF segment $s^{\star}$ by computing how much of it overlaps with $s^{\star}$. A segment is considered as a GIF segment only if the overlap is larger than 66\%. Segments without any overlap serve as negatives. The segments are then fed into our robust deep ranking model, described next.

\begin{figure}
\centering
\includegraphics[clip,trim=0 0 0 0, width=1.0\linewidth]{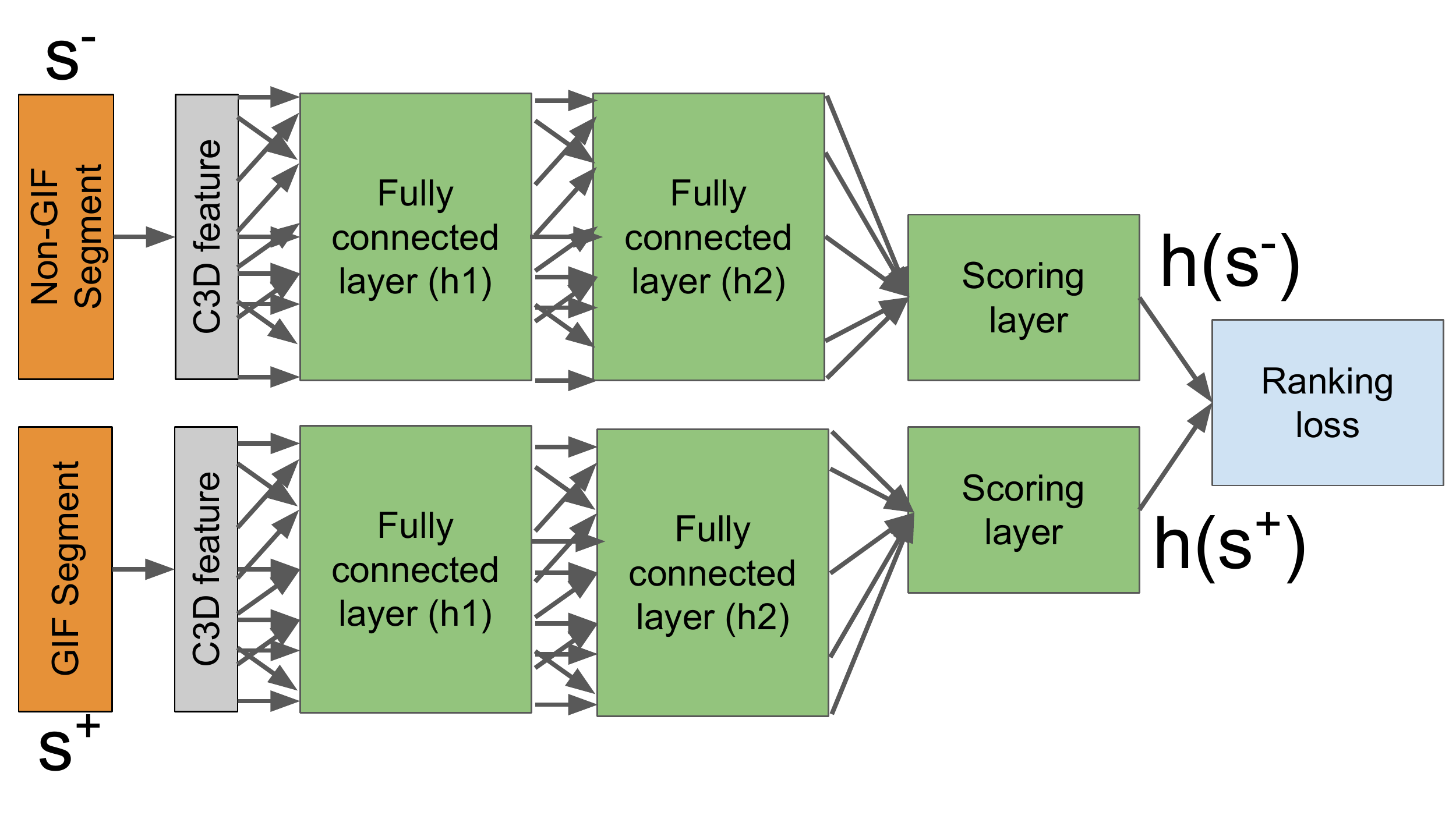}
\caption{The architecture of our Robust Deep RankNet. We train the green-colored layers from scratch. Each hidden layer is followed by a ReLu non-linearity~\cite{nair2010rectified}. The final scoring layer is a linear function of the last hidden layer. The rank loss acts on pairs of segments and is non-zero, unless $s^+$ scores higher than $s^-$ by a margin of 1. To emphasize that the loss acts on pairs of segments, we show the two passes separately, but we use a single  network.}.
\label{fig:architecture}
\end{figure}

\subsection{Robust Deep RankNet}

\noindent\textbf{Architecture overview.} Figure~\ref{fig:architecture} illustrates the architecture of our model. During training, the input is a pair of GIF and non-GIF segments. The model learns a function $h: \mathbb{R}^{d}\rightarrow\mathbb{R}$ that maps a segment $s$ to its \textit{GIF-suitability} score $h(s)$. This score is of course unknown even during training; we learn the function by comparing the training segment pairs so that a GIF segment gets a higher score than a non-GIF segment. During testing, the model is given a single segment and computes its GIF-suitability score using the learned scoring function. We compute the score $h(s)$ for all segments $s\in\mathcal{S}$ and produce a ranked list of the segments for their suitability as an animated GIF.

\smallskip\noindent\textbf{Feature representation.} Animated GIFs contain highly dynamic visual content; it is crucial to have feature representation that captures this aspect well. To capture both the spatial and the temporal dynamics of video segments, we use C3D~\cite{Tran2014a} pretrained on the Sports-1M dataset~\cite{Karpathy2014} as our feature extractor. C3D extends the image-centric network architecture of AlexNet~\cite{Krizhevsky} to the video domain by replacing the traditional 2D convolutional layers with a spatio-temporal convolutional layer, and has been shown to perform well on several video classification tasks~\cite{Tran2014a}. 

Inspired by previous methods using category specific models~\cite{Sun2014,Potapov2014}, we optionally add contextual features to the segment representation.
These can be considered meta-information, supplementing the visual features. They have the potential to disambiguate segment rankings and allow a model to score segments conditioned on the semantic category of a video.
The features include the category label, a semantic embedding of the video tags (mean over their word2vec representation~\cite{Mikolov2013a}) and positional features.
For positional features, we use the timestamp, rank and the relative position of the segment in the video.

\smallskip\noindent\textbf{Problem formulation.} A straightforward way to formulate our problem is by posing it as a classification problem, i.e., treat GIF and non-GIF segments as positive and negative examples, respectively, and build a binary classifier that separates the two classes of examples. This formulation, however, is inadequate for our problem because there is no clear cut definition of what is a good or a bad segment. Rather, there are various degrees of GIF suitability that can only be inferred by comparing GIF and non-GIF pairs.

A natural formulation is therefore posing it as a ranking problem. We can define a set of rank constraints over the dataset $\mathcal{D}$, where we require GIF segments $s^+$ to rank higher than non-GIF segments $s^-$, \ie
\begin{align*}
h(s^+) > h(s^-), \:\:\:\:
\forall \left( s^+, s^- \right) \in \mathcal{D}.
\end{align*}
This formulation compares two segments even if they are from different videos. This is problematic because a comparison of two segments is meaningful \textit{only within the context of} the video, \eg, a GIF segment in one video may not be chosen as a GIF in another. To see this, some videos contain many segments of interest (\eg compilations), while in others even the selected parts are of low quality. The notion of GIF suitability is thus most meaningful only within, but not across, the context of a single video. 

To account for this, we revise the above video-agnostic ranking formulation to be video-specific, \ie
\begin{align*}
h(s^+) > h(s^-), \:\:\:\:
\forall \left( s^+,s^- \right) \in S.
\end{align*}
That is, we require a GIF segment $s^+$ to score higher than negative segments $s^-$ that come from the same video only. Next we define how we impose the rank constraints.

\begin{figure}
\centering
\includegraphics[width=1.0\linewidth]{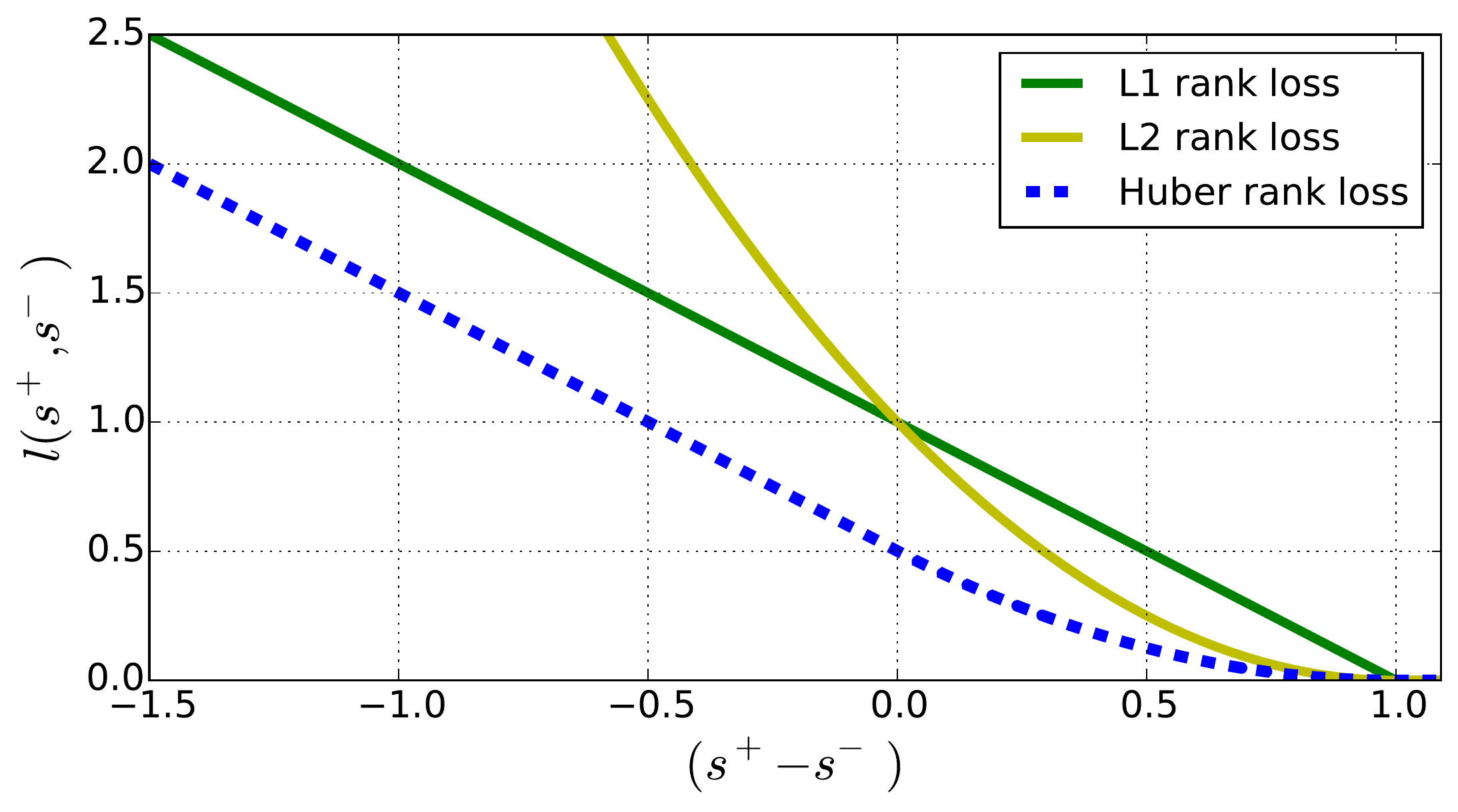}
\caption{\textbf{Rank loss comparison.} Ours Huber rank loss combines the robustness~\wrt to small margin violations of the $l_{2}$ loss with the robustness to outliers of the $l_{1}$ loss.}
\label{fig:loss}
\end{figure}

\smallskip\noindent\textbf{Loss function.}\label{sec:loss}
One possible loss function for the ranking problem is an $l_{p}$ loss, defined as
\begin{equation}
l_p(\subsup{s}{}{+},\subsup{s}{}{-})=\max\left(0,1-h(s^+)+h(s^-)\right)^{p},
\end{equation}
where $p=1$~\cite{Joachims2002} and $p=2$~\cite{Sun2014,Lee2014} are the most popular choices. The $l_{p}$ loss imposes the ranking constraint by requiring a positive segment to score higher than its negative counterpart by a margin of 1. If the margin is violated, the incurred loss is linear in the error for the $l_{1}$ loss, while for the $l_{2}$ loss it is quadratic. One drawback of the $l_{1}$ loss, compared to the $l_{2}$ loss, is that it over-penalizes small margin violations. The $l_{2}$ loss does not have such problem, but it quadratically penalizes margin violations, and thus is more affected by outliers (see Figure~\ref{fig:loss}).

Our dataset contains animated GIF contents created by online users, so some of the contents will inevitably be of low quality; these can be considered as outliers. This motivates us to propose a novel robust rank loss, which is an adaption of the Huber loss formulation~\cite{huber1964robust} to the ranking setting. This loss gives a low penalty to small violations of the margin (where the ranking is still correct), and is more robust to outliers compared to the $l_{2}$ loss. We define our loss as
\begin{equation}
\label{eq:loss}
    l_\text{Huber}(\subsup{s}{}{+},\subsup{s}{}{-}) =
\begin{cases}
   \frac{1}{2}l_2(\subsup{s}{}{+},\subsup{s}{}{-}), & \text{if } u \leq \delta \\
     \delta l_1(\subsup{s}{}{+},\subsup{s}{}{-}) -\frac{1}{2}\delta^2,                & \text{otherwise}
\end{cases}
\end{equation}
where $u=1-h(s^+)+h(s^-)$.
Thus, if the margin is violated, the loss corresponds to a Huber loss, which is squared for small margin-violations and linear for stronger violations. The parameter $\delta$ defines the point at which the loss becomes linear. We illustrate the three different forms loss functions in Figure~\ref{fig:loss}.

Considering the source of our dataset (social media), not all GIFs are expected to be of equal quality. Some might be casually created by beginners and from mediocre videos, while others are carefully selected from a high quality source. Thus, some GIFs can be considered more reliable as positive examples than others.
We take this into account by making the parameter $\delta$ GIF dependent: We assign a higher value to $\delta$ to more popular GIFs.
Our intuition behind this adaptive scoring scheme is that popular GIFs are less likely to be outliers and therefore do not require a loss that becomes linear early on.

\smallskip\noindent\textbf{Objective function.} Finally, we define our objective as the total loss over the dataset $\mathcal{D}$ and a regularization term with the squared Frobenius norm on the model weights $\mathbf{W}$:
\begin{equation}
L(\mathcal{D},\mathbf{W}) = \sum_{S_i \in \mathcal{D}}\sum_{(s^+,s^-) \in S_i}  l_\text{Huber}(\subsup{s}{}{+},\subsup{s}{}{-}) +\lambda||\mathbf{W}||^2_F,
\label{eq:loss_global}
\end{equation}
where $\lambda$ is the regularization parameter.

\subsection{Implementation Details}
We experimented with various network architectures. While the loss function turned out to be crucial, we empirically found that performance remains relatively stable for different depths of a network. Thus, we opt for a simple 2 hidden layer fully-connected model, where each hidden unit is followed by a ReLu non-linearity~\cite{nair2010rectified}. We use \ATT{512} units in the first and \ATT{128} in the second hidden layer. The final prediction layer, which outputs $h(s)$, is a simple single linear unit, predicting an unnormalized scalar score. The final network has 2,327,681 parameters. 

We minimize the objective in Eq.~\ref{eq:loss_global} using mini-batch stochastic gradient descent with backpropagation~\cite{rumelhart1985learning}. We use mini-batches of 50 pairs. In order to accelerate convergence, we apply Nesterov's Accelerated Momentum~\cite{bengio2013advances} for updating the weights. The momentum is set to 0.9 and $\lambda=0.001$ (weight decay). We initialize training with a learning rate of \ATT{$0.001$} and reduce it every 10th epoch. The learning is stopped after \ATT{25} epochs. 
We apply dropout~\cite{Hinton2014} regularization to the input (\ATT{0.8}) and after the first hidden layer (\ATT{0.25}). Dropout is a simple, approximate way to do model averaging that increases robustness of the model~\cite{Hinton2014}.

We obtain the training set segment pairs $(s^+,s^-)$ by using all positive segments, randomly sampling $k=4$ negatives per video, and combining them exhaustively. We limit the negatives in order to balance the positive-negative pairs per video. Finally, we obtain 500K pairs for training. 
For the Huber loss with a fixed $\delta$ we set $\delta=1.5$ based on the performance on the validation set. For the adaptive Huber loss, we set $\delta=1.5+p$, where p is normalized viewcount proposed in~\cite{Khosla}.

In order to further decrease the variance of our model, we use model averaging, where we train multiple models from different initializations and average their predicted scores. The models were implemented using Theano~\cite{bergstra2010theano} with Lasagne~\cite{sander_dieleman_2015_27878}.

\section{Experiments}
\label{sec:experiments}
We evaluate our method against several state-of-the art methods on our dataset. In Section~\ref{sec:highlight} we further evaluate cross-task performance on the highlight dataset of~\cite{Sun2014}.

\smallskip
\noindent
\textbf{Evaluation metrics.}
Two popular performance metrics used in video highlight detection are mean Average Precision (mAP)~\cite{Sun2014} and average meaningful summary duration (MSD)~\cite{Potapov2014}. Both mAP and MSD are, however, sensitive to video length: the longer the video is, the lower the score (think about finding the needle in the haystack). To compensate for a variety of video lengths in our dataset (see Figure~\ref{fig:video_length}), we propose a normalized version of MSD.

The normalized MSD (nMSD) corresponds to the relative length of the selected GIF at a recall rate of $\alpha$. We define it as:
\begin{equation}
nMSD=\frac{|G^*| - \alpha|G^{gt}|}{|\mathcal{V}|-\alpha|G^{gt}|},
\end{equation}
where $|.|$ denotes the length of a GIF or video, and $G^*$ is the GIF with  $\alpha$ recall \wrt the ground truth GIF $G^{gt}$. The score is normalized with the length of the ground truth GIF and the video $\mathcal{V}$, such that it is 0 if the selection equals to $G^{gt}$ (is perfect), and 1 if the ground truth has the lowest predicted score. 
The added normalization helps make the scores of different videos more comparable, in contrast to mAP, which is strongly affected by the length of the ground truth GIF, relative to the video. To account for inaccuracies in segmentation we set $\alpha=0.5$. For videos with multiple GIFs we use their mean nMSD as the video score. In addition to nMSD, we also evaluate performance using the traditional mAP.

\subsection{Compared Methods}
We compare our method to three state-of-the-art methods in highlight detection. We also provide an approximate upper bound that puts the results in perspective. 


\textbf{Domain-specific highlights~\cite{Sun2014}.} We learn a domain-specific rankSVM per video category.
Sun~\etal~\cite{Sun2014} use an EM-like approach to handle long, loosely selected highlights.
For our dataset, this problem does not occur because the GIFs are already short and focused.
We therefore simply train a rankSVM~\cite{Lee2014} per video category using C3D features. We set $C=1$ for all models.

\textbf{Deep visual-semantic embedding~\cite{Liu2015}.}
We train a network using triplets of segment, true and random titles $(s^+,t^+,t^-)$. The titles are embedded into $\mathbb{R}^{300}$ using word2vec~\cite{Mikolov2013a}.
In contrast to our method, the loss of~\cite{Liu2015} is defined over positive and negative titles and uses only positive segments (or images in their case) for training.

\textbf{Category-specific summarization~\cite{Potapov2014}.} 
This approach trains a one-vs-all SVM classifier for each video category. Thus,
the classifier learns to separate one semantic class from the others. 
At test time it uses the classifier confidence to assign each segment an importance score, which we use to obtain a ranked list.

\textbf{Approximate upper bound.}
This bound provides a reference for how well an automatic method can perform.
To obtain the upper bound, we first find all videos in our dataset that have animated GIFs from multiple creators. 
We then evaluate the performance of one GIF w.r.t. the remaining ones from the same video.
Thus, the approximate upper bound is the performance users achieve in predicting the GIFs of other users.
And it allows us to put the performance of automatic methods in perspective. We note, however, that this bound is only approximate because it is obtained in a very different setting than other methods.

\subsection{Results and Discussions}

\begin{figure*}
\centering
\includegraphics[width=1.0\linewidth]{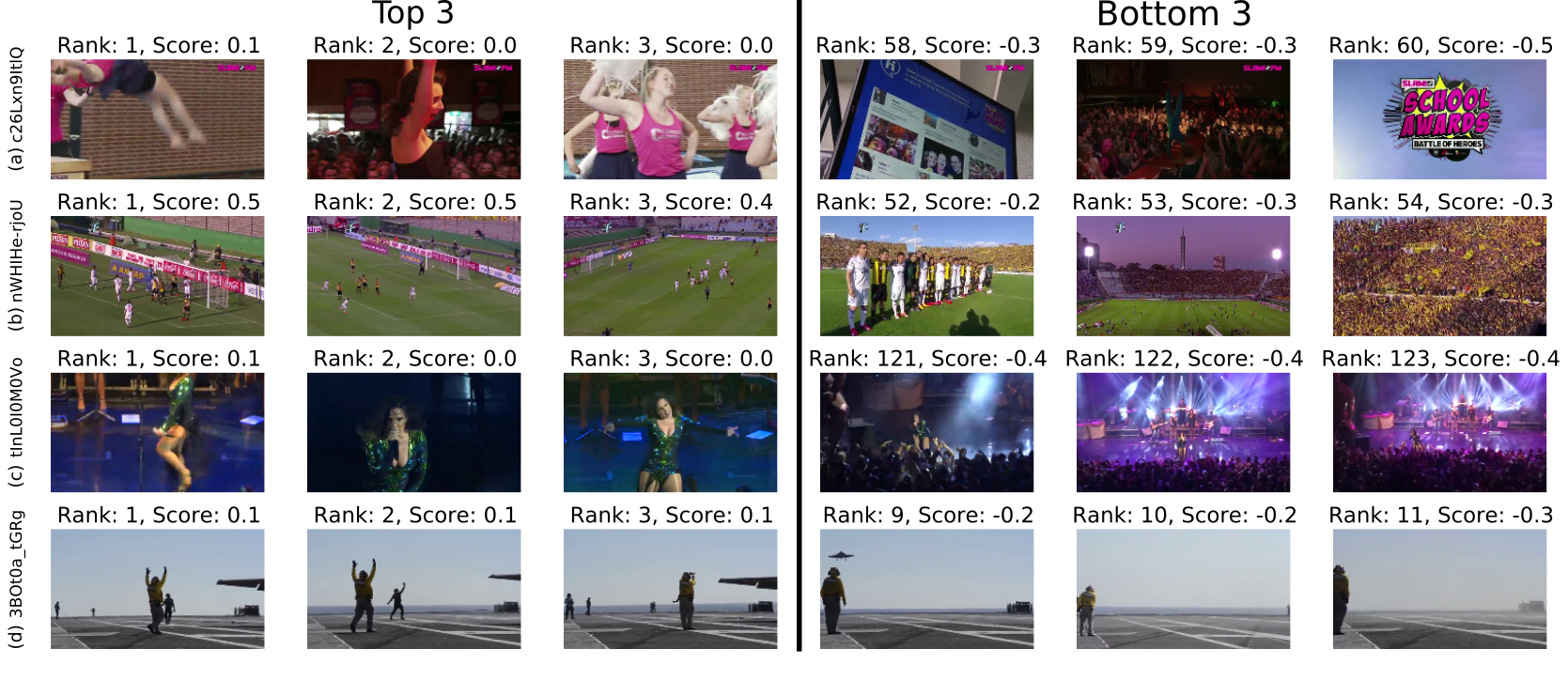}
\caption{\textbf{Qualitative results.} Examples of top 3 and bottom 3 predicted segments. Our approach picks up aspects that are related to GIF suitability. For example, it learns that segments with people in motion are suitable for GIFs (e.g., (a) and (c)), while low contrast segments without any (main) objects are not (e.g., (a) the 4th image). It also scores segments showing the goal area of soccer games higher than the crowd in the stadium (b). We show a failure case (d): the network scores the segments with people on the ground higher than the landing plane (4th image). We provide more examples in GIF format on \url{http://video2gif.info}}
\label{fig:qualitative_results}
\end{figure*}

Table~\ref{tab:test_results} summarizes the results. Figure~\ref{fig:qualitative_results} shows qualitative results obtained using our method. As can be seen, our method (``Ours'' in Table~\ref{tab:test_results}) outperforms the baseline methods by a large margin in terms of nMSD.
The strongest baseline method is domain-specific rankSVM~\cite{Sun2014}. Their learning objective is similar to ours, ~\ie, they use pairs of positive and negative segments from the same video for training. In contrast, two other baselines~\cite{Potapov2014,Liu2015} use a ``proxy'' objective, ~\ie, learn semantic similarity of segments to video category~\cite{Potapov2014} or segments to video title~\cite{Liu2015}. We believe this different training objective is crucial, allowing both our method and rankSVM~\cite{Sun2014} to significantly outperform the two baselines.

Domain-specific rankSVM~\cite{Sun2014} with C3D features performs fairly well; but our method outperforms it. 
We believe the reason for this performance difference is two-fold: (1) the $l_2$ loss in \cite{Sun2014} is not robust enough to outliers; and (2) the learning capabilities of \cite{Sun2014} are limited by the use of a linear model, compared to highly nonlinear neural nets. Next, we analyze different configurations of our method in greater detail and discuss impacts of each design choice.
The configurations differ in terms of used inputs, network architecture and objective.

\begin{table}
\centering
\begin{tabular}{|l|c|c|}
\hline
\textbf{Method} & \textbf{nMSD $\downarrow$ } & \textbf{mAP $\uparrow$} \\
\hline
\hline
Joint embedding~\cite{Liu2015} & 54.38\%  & 12.36\% \\
Category-spec. SVM~\cite{Potapov2014} & 52.98\%  & 13.46\% \\
Domain-spec. rankSVM~\cite{Sun2014} & 46.40\%  & 16.08\%  \\
\hline
\hline
Classification &  61.37\% & 11.78\% \\
Rank, video agnostic & 53.71\% & 13.25\% \\
\hline
Rank, $l_1$ loss & 44.60\% & 16.09\%	 \\
Rank, $l_2$ loss & 44.43\% & 16.10\% \\
\hline
Rank, Huber loss & 44.43\% & \textbf{16.22\%} \\
Rank, adaptive Huber loss & 44.58\% & 16.21\% \\
\hline
\hline
 \parbox{5cm}{Rank, adaptive Huber loss \\ + context (Ours)} & 44.19\%  & 16.18\% \\
\hline
Ours + model averaging & \textbf{44.08\%} & 16.21\% \\
\hline
\hline
Approx. bounds & 38.77\%  & 21.30\% \\
\hline
\end{tabular}
\caption{Experimental results. A lower nMSD and higher mAP represent better performance.}
\label{tab:test_results}
\end{table}

\smallskip
\noindent
\textbf{What loss function is most robust?}
We analyze performance with different loss functions and training objectives discussed in Section~\ref{sec:loss}.
As expected, classification models always performs poorly compared to ranking models. Also, using video agnostic training data performs poorly. This indicates that the definition of a highlight is most meaningful within the video. 
When comparing $l_1$ loss and $l_2$ loss, we find that $l_1$ loss penalizes small margin violations (\ie, $0 < h(s^{+})-h(s^{-}) < 1$) too strongly, while the $l_2$ loss is affected by outliers. Our Huber rank loss avoids the two issues by combining the robustness to outliers ($l_1$ loss) and the decrease in the gradient for small margin violations ($l_2$ loss); it thus performs better than the other losses.

\smallskip
\noindent
\textbf{The role of context.}
Inspired by previous methods using category specific models~\cite{Sun2014,Potapov2014} we used contextual information as input to our model (category label, a semantic  of the video tags and positional features).
When comparing the performance with and without context, we find that they perform similarly (Table~\ref{tab:test_results}). 
We believe that most of the information about the context is already present in the segment representation itself.
This is supported by~\cite{LinICCV15} who show that the context can be extracted from the segment itself with high accuracy.

\subsection{Cross Dataset Performance}
\label{sec:highlight}
As discussed, automatic GIF creation is related to video highlight detection.
Of course, they are not identical: GIFs have a different focus and often depict funny or emotional content rather than highlights only. Nonetheless, it is interesting to see how well our method generalizes to this task. 

We evaluate our model on the dataset of~\cite{Sun2014}, which contains videos from hand-selected categories such as surfing and skiing. We also evaluate the best performing baseline, domain-specific rankSVM~\cite{Sun2014}, trained on our dataset and tested on the highlight dataset. The results are summarized in  Table~\ref{tab:highlight_results} (we borrow previously reported results~\cite{Yang2015,Sun2014}). 

Our method outperforms rankSVM by a large margin, which suggests that our model generalizes much better than the baseline method. It also significantly outperforms the method of Yang~\etal~\cite{Yang2015}, which trains an auto-encoder model for each domain. 
Sun~\etal~\cite{Sun2014} tops the performance, but they use video category labels (which are handpicked) and learn multiple models, one per category, directly on the highlight dataset. Instead, our method learns a single global model on the GIF data, with much more diverse video categories. Nonetheless, it shows competitive performance.


\begin{table}
\centering
\begin{tabular}{|l|l|l|l|l|l|}
\hline
\textbf{Category} & Ours  & {\small rankSVM} & Yang~\cite{Yang2015}  & Sun~\cite{Sun2014}\\
\hline
             skating & 55.4\% & 26.2\% & 25\% & 61\% \\
          gymnastics & 33.5\% & 25.5\% & 35\% & 41\%  \\
             surfing & 54.1\% & 45.0\% & 49\% & 61\%  \\ 
                 dog & 30.8\% & 47.3\% & 49\% & 60\%  \\
             parkour & 54.0\% & 44.7\% & 50\% & 61\%  \\
              skiing & 32.8\% & 35.6\% & 22\% & 36\%  \\
              \hline\hline
              Total  & 46.4\% & 37.9\% & 41.2\% & 53.6\%\\
\hline
\end{tabular}
\caption{Cross-dataset results (mAP). We train on our dataset and test on the video highlight dataset of~\cite{Sun2014}. Our method outperforms rankSVM and \cite{Yang2015}, which learns an unsupervised model for each domain. Sun~\etal~\cite{Sun2014} performs best, but it is directly trained on their dataset and learns multiple models, one per category. Instead, we learn a single global model for GIF suitability.}
\label{tab:highlight_results}
\end{table}

\section{Conclusion}
We introduced the problem of automatically generating animated GIFs from video, and proposed a Robust Deep RankNet that predicts the GIF suitability of video segments. Our approach handles noisy web data with a novel adaptive Huber rank loss, which has the advantage of being robust to outliers and able to encode the notion of content quality directly into the loss. On our new dataset of animated GIFs we showed that our method successfully learns to rank segments with subtle differences, outperforming existing methods. Furthermore, it generalizes well to highlight detection.

Our novel Video2GIF task, along with our new large-scale dataset, opens the way for future research in the direction of automatic GIF creation. For example, more sophisticated language models could be applied to leverage video meta data, as not all tags are informative. Thus, we believe learning an embedding specifically for video tags may improve a contextual model.
While this work focused on obtaining a meaningful ranking for GIFs, we only considered single segments.
Since some GIFs range over multiple shots, it would also be interesting to look at when to combine segments or even do joint segmentation and selection.

{\small
\bibliographystyle{ieee}
\bibliography{video_sum,interest,general}
}

\end{document}